\documentclass[wcp]{jmlr}

\usepackage{longtable}

\usepackage{booktabs}

\usepackage{lineno}

\makeatletter
\let\Ginclude@graphics\@org@Ginclude@graphics 
\makeatother

\jmlrvolume{222}
\jmlryear{2023}
\jmlrworkshop{ACML 2023}

\title[Emphasizing Main Concepts over Foreground Influence]{Mitigating  Bias: Enhancing Image Classification by Improving Model Explanations}

 \author{\Name{Raha Ahmadi*} \Email{rahaahmadi@aut.ac.ir}\\
  \Name{Mohammad Javad Rajabi*} \Email{rajabi2001@aut.ac.ir}\\
  \Name{Mohammad khalooei} \Email{khalooei@aut.ac.ir}\\
  \addr Amirkabir University of Technology\\
  \Name{Mohammad Sabokrou} \Email{mohammad.sabokrou@oist.jp}\\
  \addr   Okinawa Institute of Science and Technology \\Institute for Research in Fundamental Sciences(IPM)}

\editors{Berrin Yan{\i}ko\u{g}lu and Wray Buntine}

\begin{document}

\maketitle

\def\thefootnote{*}\footnotetext{Equal contribution}

\begin{abstract}

Deep learning models have demonstrated remarkable capabilities in learning complex patterns and concepts from training data. However, recent findings indicate that these models tend to rely heavily on simple and easily discernible features present in the background of images, rather than the main concepts or objects they are intended to classify. This phenomenon poses a challenge to image classifiers as the crucial elements of interest in images may be overshadowed. In this paper, we propose a novel approach to address this issue and improve the learning of main concepts by image classifiers. Our central idea revolves around concurrently guiding the model's attention toward the foreground during the classification task. By emphasizing the foreground, which encapsulates the primary objects of interest, we aim to shift the focus of the model away from the dominant influence of the background. To accomplish this, we introduce a mechanism that encourages the model to allocate sufficient attention to the foreground. We investigate various strategies, including modifying the loss function or incorporating additional architectural components, to enable the classifier to effectively capture the primary concept within an image. Additionally, we explore the impact of different foreground attention mechanisms on model performance and provide insights into their effectiveness. Through extensive experimentation on benchmark datasets, we demonstrate the efficacy of our proposed approach in improving the classification accuracy of image classifiers. Our findings highlight the importance of foreground attention in enhancing model understanding and representation of the main concepts within images. The results of this study contribute to advancing the field of image classification and provide valuable insights for developing more robust and accurate deep-learning models.
\end{abstract}

\section{Introduction}
\begin{figure}[!t]
\begin{center}
\includegraphics[width=0.6\textwidth]{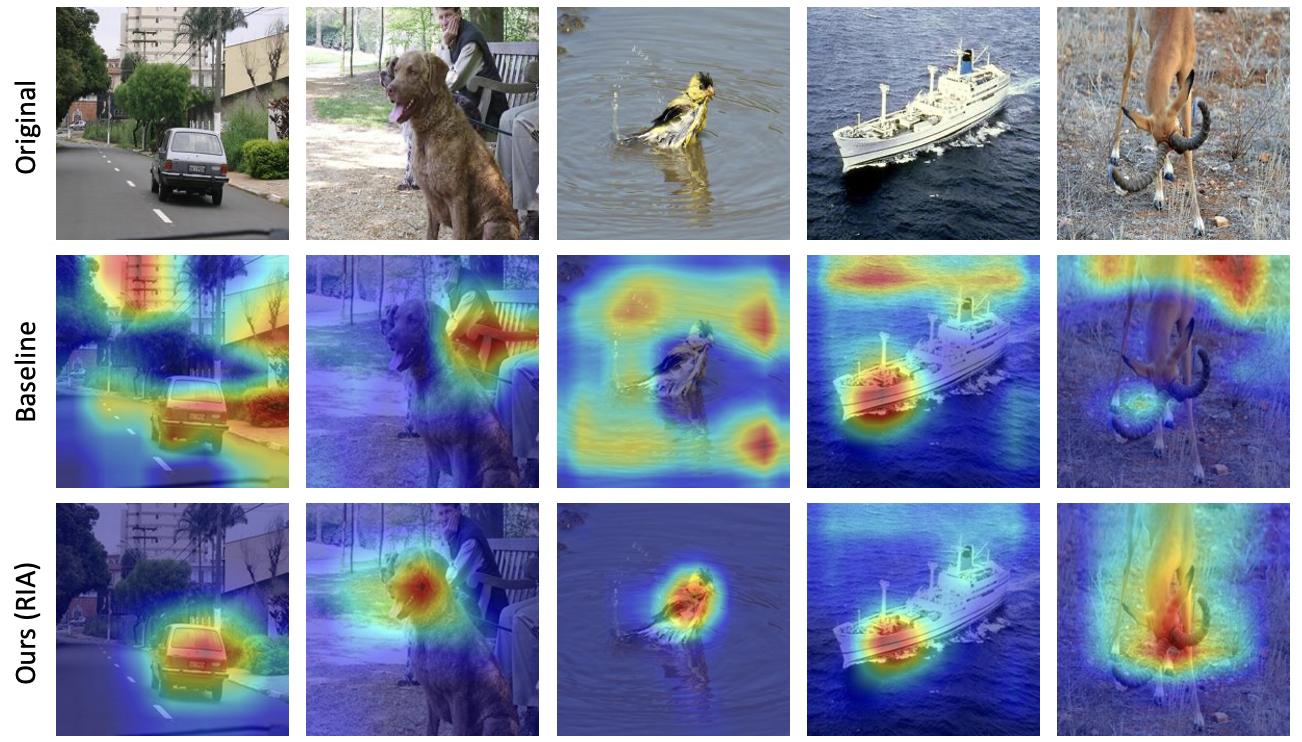}
\caption{Our approach demonstrates enhanced robustness of the Grad-CAM explanation and reduced attention to background regions, resulting in more accurate and focused explanations.}\label{fig:intro-fig}
\vspace{-\baselineskip}
\end{center}
\end{figure}

Deep neural networks (DNNs) have gained widespread adoption in various computer vision tasks due to their superior performance and remarkable capabilities in learning complex patterns and concepts from training data. Nevertheless, recent studies have revealed that DNNs generally learn the most uncomplicated features, like biases, and do not concentrate well on the main key objects of input image \cite{mehmanchi2023revealing, singh2020don}. Additionally, most of the computer vision classification tasks experienced similar issues, where the models were biased towards the non-central objects or the background of the primary object in the input sample.  \cite{moayeri2022comprehensive} has discovered that highly precise models can be greatly affected by background noise, more so than foreground noise. This indicates that in complex scenarios, deep learning models unexpectedly depend on non-central objects or features in the background rather than accurately identifying the primary key objects that were programmed to be classified. This poses a significant challenge to the reliability of DNNs.


In general, the interpretability of DNNs has thus emerged as an important research area, with the purpose of enhancing  transparency and applicability of DNNs. This makes it possible to provide trustfully and helps identify any spurious correlations the network may have inadvertently learned to use to make its decision \cite{singh2020don}. The use of Explainable Artificial Intelligence (XAI) approaches can be considered valuable for interpreting DNNs and pointing out biases. It is unfortunate that methods like Grad-CAM \cite{selvaraju2017grad}, have proven to be unreliable. For instance, when a cat image and its rotated version are given to the classifier; it recognizes them as a cat, but their explainability results from Grad-CAM differ. \cite{pillai2022consistent} raises this concern and proposes a training method inspired by contrastive self-supervised learning to address this issue.

To address challenges and prevent bias toward the background in image classification, We propose aligning the Grad-CAM interpretations mechanism with the main objects of interest. Inspired by self-supervised learning, our method uses a novel loss function to align the attention area of the model with the main object. This approach aims to enhance trustworthiness, reduce sensitivity to background noise, and improve the model's interpretability in complex computer vision tasks.

We evaluated our proposed Region of Interest Activation (RIA) method to assess its impact on classification accuracy and the sensitivity of the model to the foreground and background attributes. Our method aims to guide the model in prioritizing the most discriminative features of the input image, thereby enhancing accuracy and reliability in fine-grained classification scenarios.

Figure \ref{fig:intro-fig} illustrates some examples which demonstrate that the RIA method significantly improves the interpretation of the model in primary objects of corresponding input images, as identified by an object detector. Furthermore, we observed that our method enhances classification accuracy in fine-grained settings and when faced with foreground and background noise.
The evaluation highlights the effectiveness of our RIA method in improving the accuracy and reliability of deep learning models for image classification tasks. By directing models to focus on relevant features and reducing sensitivity to irrelevant attributes, our method has the potential to enhance performance across various applications reliant on image classification.

Our main contributions are summarized as follows:

 \begin{enumerate}
  \item Our study highlights the significance of how models can be biased towards objects or features in the background, which can impact the model's explanations and lead to inaccurate decisions.
  \item We propose a novel approach that promotes the model's focus on the primary object of the image by utilizing the proximity between the model's attention area and the detected area of an object.
  \item We have adapted the basic IoU loss to account for the anticipated inaccuracies in estimating the area of an object by an object detector.
  \item Our approach not only enhances the accuracy of the model but also significantly improves its reliability and additionally would be more robust under the foreground and background noises.
\end{enumerate}

\section{Related Works}
\subsection{Interpretability Methods}
Interpretability methods for deep neural networks have been extensively studied in recent years, driven by the widespread adoption of these models across various tasks. \cite{ribeiro2016should} have highlighted that machine learning models often capture undesirable correlation artifacts during training, which can be challenging to identify solely by relying on prediction accuracy. In order to address this issue, several methods have been proposed to detect salient regions in images. \cite{zeiler2014visualizing} introduced an approach that leveraged gradients of the class conditional output with respect to the input image. By identifying spatial locations with large gradient magnitudes, a saliency map corresponding to the class could be obtained. Building upon this work, \cite{springenberg2014striving} and \cite{sundararajan2017axiomatic} further enhanced the quality of saliency maps, resulting in sharper visualizations. \cite{ross2017right} showed that constraining the gradient explanations to be small in irrelevant areas using an annotation mask improved the quality of these explanations, albeit with additional computational costs. Class Activation Mapping (CAM) was introduced by \cite{zhou2016learning} to produce a coarse localization heatmap by utilizing a global average pooling (GAP) layer to calculate the gradients flowing into the final convolution layer. Gradient-weighted Class Activation Mapping (Grad-CAM), proposed by \cite{selvaraju2017grad}, extends the concept of CAM by utilizing gradients flowing into the final convolutional layer for a specific class. This approach generates a coarse localization map that highlights the important regions in the image for predicting the corresponding class. 

Our proposed method is also based on Grad-CAM heatmaps. However, we introduce a novel approach by using Grad-CAM during the training process. This allows us to guide the generation of heatmaps, resulting in improved explanations that align more accurately with the known regions describing the desired features.

\subsection{Explanation-guided learning}
In recent studies, researchers have explored the integration of explanations during model training to enhance predictive performance. \cite{rieger2020interpretations} aims to align explanations with human annotations based on domain knowledge. This strategy helps reduce the model's reliance on background pixels. However, obtaining ground truth explanations can be labor-intensive \cite{wang2020self} or even unfeasible due to subjectivity in real-world tasks \cite{roscher2020explainable}. \cite{pham2021learning} employed segmentation masks to direct attention maps towards important regions of images for attribute prediction tasks. \cite{selvaraju2021casting} utilized saliency maps generated with DeepUSPS \cite{nguyen2019deepusps} during training to guide attention maps, aiming to enhance self-supervised representation learning. Similarly, \cite{pillai2022consistent} focused on maintaining consistent explanations to facilitate generic representation learning through the use of contrastive objectives. The goal of such approaches is not only to improve performance but also to make sure that the model
is “right for the right reasons” \cite{ross2017right}. For classifiers, this typically involves jointly optimizing both classification performance and localization to object features.

In this work, we present a novel loss function that guides DNNs toward accurate object localization and enhances their robustness by reducing reliance on spurious features and background information.

\section{Proposed Method}
\begin{figure*}[!t]
\begin{center}
\includegraphics[width=1\textwidth]{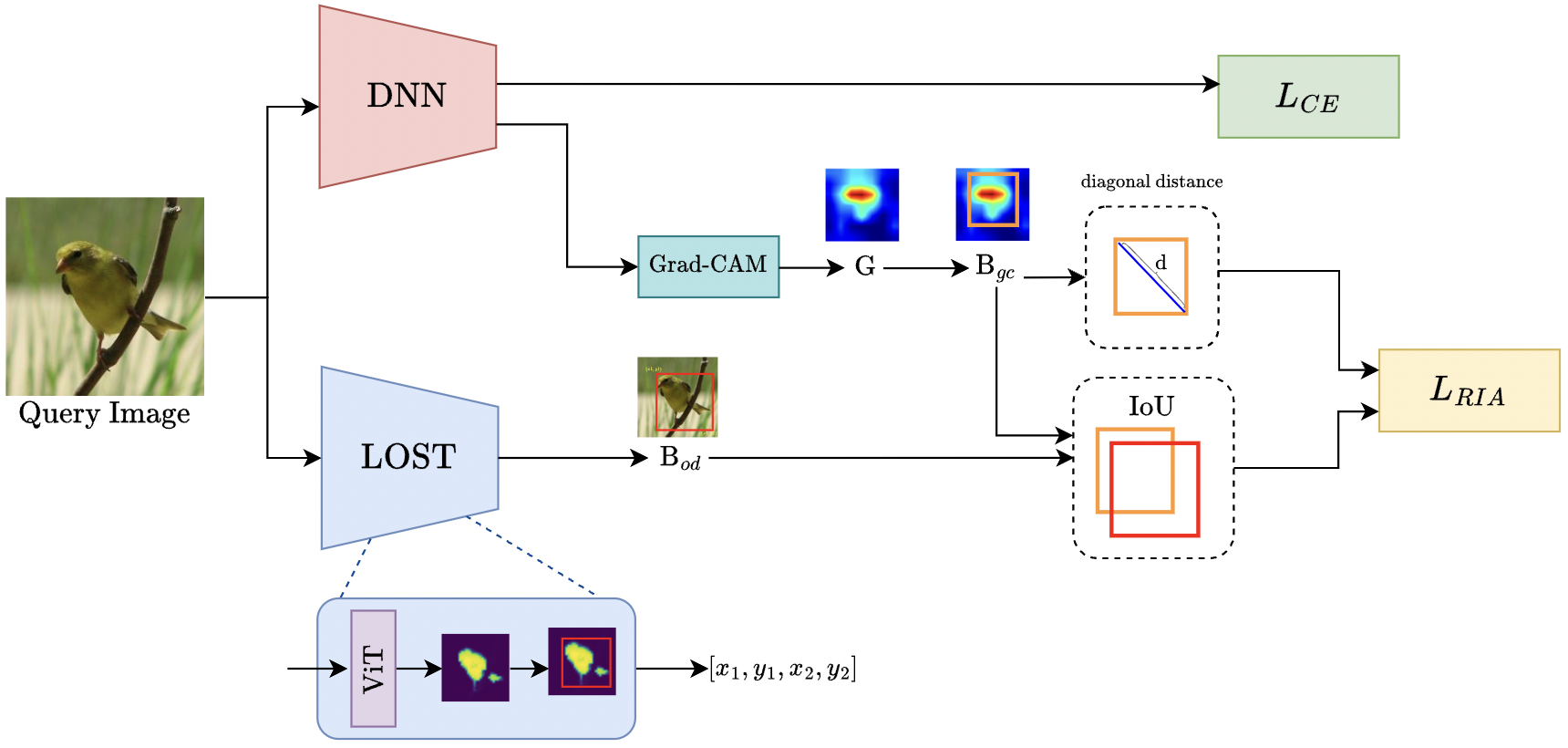}
\caption{\textbf{The block diagram of our method.} Our method consists of both cross-entropy loss $L_{CE}$ and Region of Interest Activation loss $L_{RIA}$ which represents a modified IoU loss. We load a batch of images and consider one to be the query image. We feed the query image to the network and calculate $L_{CE}$. We calculate Grad-CAM for this image on the top predicted category and obtain a bounding box from the Grad-CAM heatmap. We feed the query image to the object detector network and get a predicted bounding box. We then compute the loss between these two boxes $L_{RIA}$ and combine it with $L_{CE}$ loss.}\label{fig:diagram}
\vspace{-\baselineskip}
\end{center}
\end{figure*}

The main goal of our approach is to ensure fairness and reduce biases in image classification while maintaining the consistent interpretability of the model.  To achieve this, we propose the Region of Interest Activation Loss ($RIA$), which encourages the model to classify images accurately and focus on the main concept or object within the images. Our proposed method consists of two key components: (1) Categorical Cross-Entropy Loss ($L_{CE}$), and (2) $RIA$: Region of Interest Activation Loss (RIA). These components work together to enhance the network's ability to attend to both foreground and background objects.
To begin the learning process, we start with the $L_{CE}$ loss function, which is a standard approach for classification tasks. This loss function guides the network to minimize the discrepancy between predicted and ground truth labels.
However, to ensure that the model attends to the foreground objects as well as the background, we introduce the $RIA$ loss function. This loss function leverages the concept of Grad-CAM (Gradient-weighted Class Activation Mapping) and aims to maximize the intersection over the union of the foreground object and the Grad-CAM during training. By incorporating $RIA$, we encourage the network to focus on relevant regions of interest and improve its ability to distinguish foreground and background objects.
To combine these losses effectively, we define a new objective function and learning scenario. This combined objective function encompasses both the $L_{CE}$ loss and the $RIA$ loss, allowing the model to optimize for classification accuracy while also attending to important regions in the input data. The learning scenario incorporates these losses throughout the training process, enabling the model to learn the necessary representations. It is worth mentioning that, we exploited an unsupervised and low-cost object detector.

Figure \ref{fig:diagram} illustrates the block diagram of our method. This section provides a brief overview of the Grad-CAM interpretation algorithm and the object detection algorithm. We then delve into the details of the Region of Interest Activation Loss term.

\subsection{Background on Grad-CAM}

Grad-CAM uses the gradient information flowing into the
last convolutional layer of the CNN to assign importance
values to each neuron for a particular decision of interest. \cite{selvaraju2017grad}
To apply Grad-CAM to an input image $x$ and a deep neural network $f$, we start by obtaining the output logits y for each category by feeding $x$ to the model, where $y^{t}$ corresponds to the output for category t, and feeding y through a SoftMax operator produces the probability distribution over categories.
We then select last convolutional layer of the network and compute the derivative of the predicted output with respect to each channel of the convolutional layer, averaged over all spatial locations to get the importance of each channel of the convolutional layer in making the current prediction:

\begin{equation}
\centering
\alpha_{k}^{t} = \frac{1}{Z}\sum_{i}\sum_{j}\frac{\partial{y^{t}}}{\partial{A_{ij}^{k}}}
\end{equation}
 where $A_{ij}^{k}$ is the activations of the convolutional layer at channel k and location $(i, j)$, $Z$ is a normalizer, and $\alpha_{k}^{t}$ is the importance weights of channel $k$. Then, we perform a weighted combination of forward activation maps for each channel to get a 2D matrix over spatial locations, and follow it by a ReLU to discard negative values.

\begin{equation}
\centering
gradcam_{ij}^{t} = Relu(\sum{k}\alpha_{k}^{t}A_{ij}^{k})
\end{equation}
Finally, we resize it using bilinear interpolation to the size of the input image to get the interpretation heatmap.

\subsection{Obtaining Bounding Box from heatmap}
To obtain a bounding box ($B_{gc}$) from the heatmap generated by Grad-CAM ($G$), we apply threshold segmentation ($T$) on G as output of Grad-CAM to create Grad-CAM binary mask (GBM), where regions of interest are identified based on intensity values above the threshold.

\begin{equation}
GBM(x, y) = \begin{cases}
1, & \text{if } \text{G}(x, y) > \text{T} \\
0, & \text{otherwise}
\end{cases}
\end{equation}

Then, we have employed  connected component analysis to identify individual connected regions within the binary mask. Each connected region corresponds to a potential object.
For each connected component, we calculate the bounding box by determining the minimum and maximum coordinates in both the horizontal and vertical directions. These coordinates specify the rectangular region that tightly encapsulates the object of interest.
By following these steps, we can effectively extract the bounding box associated with the highest score in the heatmap. This bounding box allows us to precisely locate and highlight the region in the input image that played a crucial role in the neural network's classification decision.

\subsection{Unsupervised Object Detector}
To obtain the target bounding boxes, we use LOST (i.e., Localizing Objects with Self-Supervised Transformers) ~\cite{simeoni2021localizing} which can unsupervisedly detect the objects.
The LOST method utilizes self-supervised learning techniques, specifically self-supervised transformers, to train a model to discover and localize objects within an image. By localizing objects in image collections without supervision, we can improve the model's ability to become object-agnostic.

In the LOST approach, the image is divided into equal-sized patches and fed into a transformer model. Instead of focusing on the CLS token (i.e., An additional, learned vector called the class token), the key component of the last attention layer is used for computing similarities between different patches. By doing so, we can localize a part of an object by selecting the patch with the least number of similar patches, which is referred to as the seed. The justification for this seed selection criterion is based on the empirical observation that patches of foreground objects are less correlated than patches corresponding to the background. Then the seed is expanded by adding other patches that are highly correlated to it and are thus likely to be part of the same object, a process which is called seed expansion. Finally, a binary object segmentation mask is constructed by computing the similarities of each image patch to the selected seed patches. Then the bounding box of an object is considered as the box that tightly encloses the largest connected component in this mask that contains the initial seed.

\subsection{$RIA$: Region of Interest Activation Loss}
To enhance the model’s consistency and interpretability, we propose a modified Intersection over Union (IoU) loss that bridges the gap between the object detector's bounding box predictions and the bounding boxes obtained from Grad-CAM. 

Generally, the IoU-based loss can be defined as
\begin{equation}
\centering
\ IoU_{Loss} = 1 - IoU + R(B, B_{gt})
\end{equation}
where R(B, $B_{gt}$) is the penalty term for predicted box B and target box $B_{gt}$ \cite{zheng2020diou}.

IOU is a commonly used evaluation metric in computer vision and object detection tasks. It is used to measure the similarity between two arbitrary shapes (boxes).
    Generally, the IoU metric is defined as
\begin{equation}
\centering
IoU = \frac{|B \cap B_{gt}|}{|B \cup B_{gt}|}
\end{equation}

In our approach, we use the bounding box attained from Grad-CAM as the predicted box($B_{gc}$) and the bounding box generated by the object detector as the target box($B_{od}$). Our goal is to minimize the differences between the predicted and target boxes, making them as similar as possible. However,  the generated object detector boxes as the target boxes may not be entirely accurate and flawless which is not desired. These bounding boxes may be larger than necessary, covering not only the main object but also including some of the background objects. Therefore, it is important to acknowledge that these bounding boxes are not entirely error-free and may require further refinement.
To address this issue, we are considering a modification to the IoU term. Specifically, our IoU use only the box $B_{gc}$ in the denominator instead of the union of $B_{gc}$ and $B_{od}$.
\begin{equation}
\centering
\hat{IoU} = \frac{|B_{od} \cap B_{gc}|}{|B_{gc}|}
\end{equation}

This term is designed to increase the size of the intersection area without increasing the size of $B_{gc}$, thereby making the predicted bounding box equal to or smaller than the target bounding box. Additionally, this term promotes the containment of $B_{gc}$ within $B_{od}$. As a result, the numerator and denominator of $\hat{IoU}$ are identical when the intersection between the predicted($B_{gc}$) and target bounding boxes($B_{od}$) is equal to $B_{gc}$ itself.  We have found that this modification adequately compensates for any inaccuracy of the object detector, since the intersection area between the $B_{gc}$ and $B_{od}$ usually covers the main object. Our model seeks to ensure that the intersection area captures the most significant part of the main object, while excluding any irrelevant objects or background areas.

The size of the Grad-CAM box plays a crucial role in model performance. Large attention regions may encompass unnecessary features, leading to inaccurate predictions. It is important to encourage the model to focus on the most relevant regions by constraining the size of Grad-CAM boxes.
To address the issue of large attention regions and improve model performance, we propose a penalty-based modification to the loss function. This modification encourages the model to minimize the size of its attention region, resulting in smaller Grad-CAM boxes that focus on the most discriminative features.

We introduce a diagonal distance penalty for the Grad-CAM box which refers to the length of the diagonal line that spans across the box. It is calculated by measuring the Euclidean distance between the opposite corners of the box. The diagonal distance provides an estimate of the size and extent of the Grad-CAM box in terms of its spatial coverage. By this penalty, our objective is to minimize the size of the predicted bounding boxes while ensuring that they still cover the most significant part of the main object. By doing so, we aim to mitigate the influence of irrelevant objects or background areas on the model's explainability and accuracy.

\begin{equation}
\centering
R(B_{gc}) = \lambda * diagonal\ distance(B_{gc})
\end{equation}
where $\lambda$ is the regularization coefficient. With the modifications to the IoU loss and the inclusion of the diagonal distance penalty term, we have now finalized our Interest Activation loss ($L_{RIA}$) as:
\begin{equation}
\centering
L_{RIA} = 1 - \hat{IoU} + R(B_{gc})
\end{equation}

Our final loss is the combination of the standard cross-entropy loss ($L_{CE}$) and our Region of Interest Activation loss ($L_{RIA}$). Hence we minimize the following loss function:
\begin{equation}
\centering
L = \alpha * L_{CE} + \beta * L_{RIA}
\end{equation}
where $\alpha$ and $\beta$ are hyper-parameters that control the trade-off between the two loss terms.

\section{Experiment Results}
In this section, we present the results of a series of experiments conducted using our proposed method on variant models. The main objective of these experiments is to evaluate the impact of our Region of Interest Activation ($RIA$) loss on the overall performance and robustness of our models, particularly under the addition of background and foreground noise.
For each experiment, we conducted a comparative analysis between two models: the baseline model, trained solely with the standard cross-entropy loss, and the consistent model, trained with both the standard cross-entropy loss and the $RIA$ loss. By comparing the performance of these two models, as shown in Table \ref{tab:scratch_comparison} and \ref{tab:pretrained_comparison}, we can effectively assess the effectiveness of our proposed method in improving classification accuracy and model trustability.

\subsection{Dataset}
We use RIVAL10 dataset \cite{moayeri2022comprehensive}, whose samples include RIch Visual Attributions with Localization. RIVAL10 consists of images from 20 categories of ImageNet-1k \cite{deng2009imagenet}, with a total of 26k high resolution images organized into 10 classes, matching those
of CIFAR10.

\subsection{Implementation Details}
We use PyTorch \cite{paszke2019pytorch} to train and evaluate our models for all experiments.
We use pretrained Resnet18 and Resnet50 models \cite{he2016deep} with the settings used in a previous publication \cite{moayeri2022comprehensive}. Additionally, we trained VGG16 \cite{simonyan2014very}, Resnet18, and Resnet50 models from scratch. To train our models from scratch on the RIVAL10 dataset, we use a two-step training procedure. Initially, we train the models for 10 epochs without the RIA loss, allowing them to gain preliminary insights into the images. Subsequently, we incorporate the RIA loss and continue training to further refine the models using the guidance provided by the loss function. For training our models from scratch on the RIVAL10 dataset, we use Adam optimizer with a learning rate of 0.001 for Resnets and SGD with a learning rate of 0.01 for VGG16. By hyperparameter tuning, we set $\alpha$ = 1, $\beta$ = 0.5, $\lambda$ = 0.1, and $T$ = 0.5 for all experiments using our method. Models were trained on an Nvidia RTX
3090 GPU over 50 epochs.

\begin{table}[!h]
  \centering
  \begin{tabular}{|c|c|c|}
    \hline
    \textbf{Model} & \textbf{Baseline Acc (\%)} & \textbf{Ours (RIA) Acc (\%)} \\
    \hline
    VGG16 & 86.6\% & \textbf{91.7\%} \\
    \hline
    ResNet50 & 88.46\% & \textbf{88.99\%} \\
    \hline
    ResNet18 & 88.51\% & \textbf{88.78\%} \\
    \hline
  \end{tabular}
  \caption{Classification Accuracy for models trained from scratch on RIVAL10 validation set.}
  \vspace{-\baselineskip}
  \label{tab:scratch_comparison}
\end{table}

\begin{table}[!h]
  \centering
  \begin{tabular}{|c|c|c|}
    \hline
    \textbf{Model} & \textbf{Baseline Acc (\%)} & \textbf{Ours (RIA) Acc (\%)} \\
    \hline
    ResNet50 & 88.46\% & \textbf{88.99\%} \\
    \hline
    ResNet18 & 98.8\% & \textbf{99\%} \\
    \hline
  \end{tabular}
  \caption{Classification Accuracy for pretrained models on       RIVAL10 validation set.}
  \label{tab:pretrained_comparison}
  \vspace{-\baselineskip}
\end{table}

\subsection{Evaluating the Sensitivity to Background/Foreground}
\begin{figure}[!ht]
\begin{center}
\includegraphics[width=0.9\textwidth]{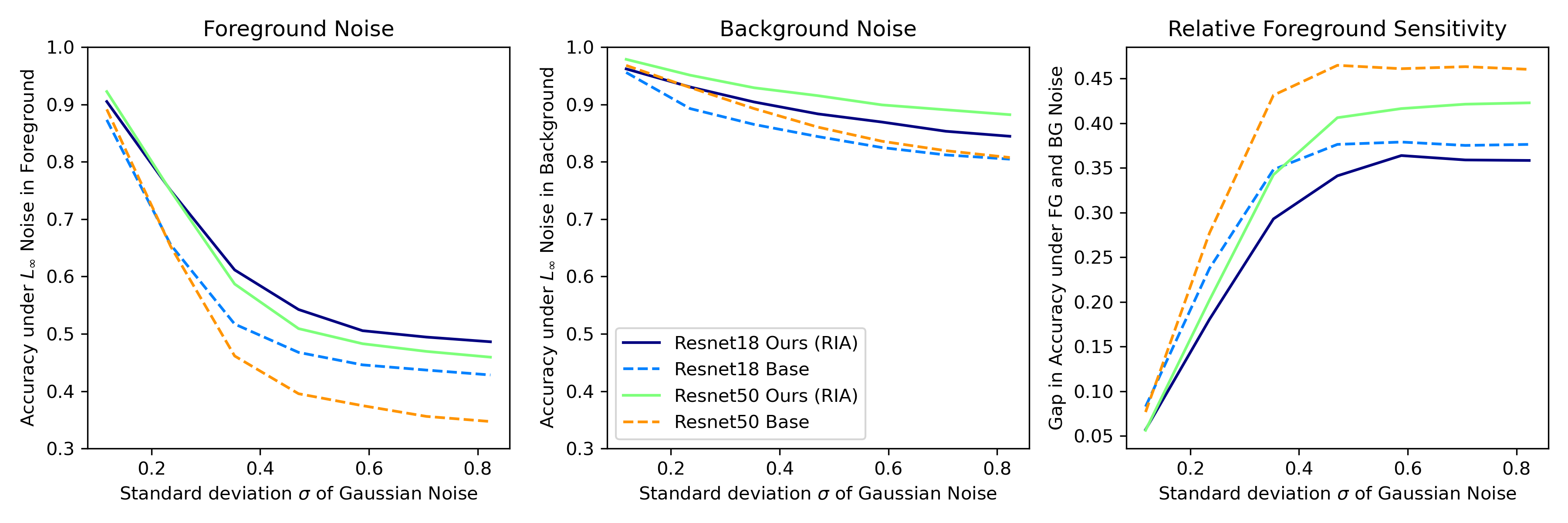}
\caption{The chart on the left displays accuracy levels under foreground (left) and background (middle) noise at different levels. The models are categorized by their architecture and training method for pretrained models, and each curve represents the average accuracy of all models in the group. On the right, the chart shows the RFS by group.}\label{fig:noise-pretrained}
\vspace{-\baselineskip}
\end{center}
\end{figure}

\begin{figure}[htp]
\begin{center}
\includegraphics[width=0.9\textwidth]{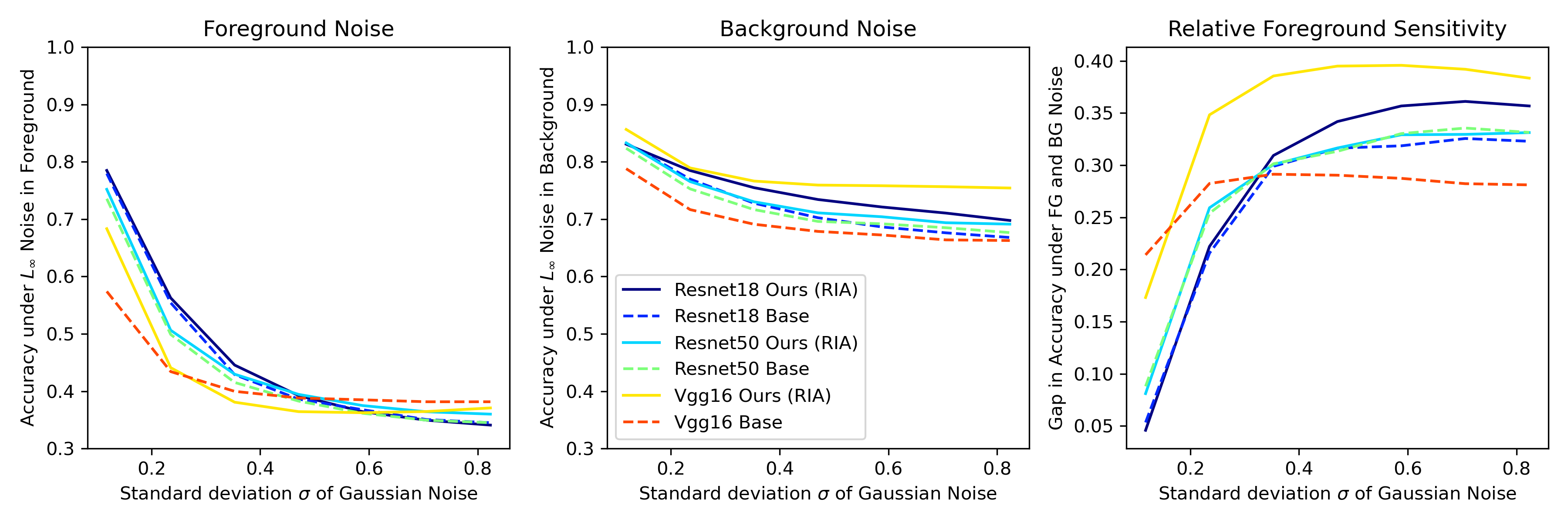}
\caption{The chart on the left displays accuracy levels under foreground (left) and background (middle) noise at different levels. The models are categorized by their architecture and training for models trained from scratch, and each curve represents the average accuracy of all models in the group. On the right, the chart shows the RFS by group.}\label{fig:noise-scratch}
\vspace{-\baselineskip}
\end{center}
\end{figure}

To assess the robustness of our method, we conducted a thorough analysis by adding Gaussian noise to both the foreground and background regions separately. This allowed us to evaluate how the corruption of each region affects the performance of our models. The evaluation of sensitivity to background and foreground in both pretrained models and models trained from scratch, as shown in Figure \ref{fig:noise-pretrained} and \ref{fig:noise-scratch}, indicates that our method outperforms the baseline approach in noisy conditions. The use of RIA loss enables us to mitigate biases towards the background and enhance foreground attention. Consequently, our models exhibit a smaller decrease in classification accuracy compared to the baseline models.

To quantify the sensitivity of a model to foregrounds relative to its sensitivity to backgrounds, we introduce relative foreground sensitivity (RFS). \cite{moayeri2022comprehensive} Let $a_{fg}$ and $a_{bg}$ denote accuracy under noise in the foreground and background, respectively. 
We then define RFS as
\begin{equation}
\centering
RFS = a_{bg} - a_{fg}
\end{equation}


\begin{figure}[!ht]
\begin{center}
\includegraphics[width=0.8\textwidth]{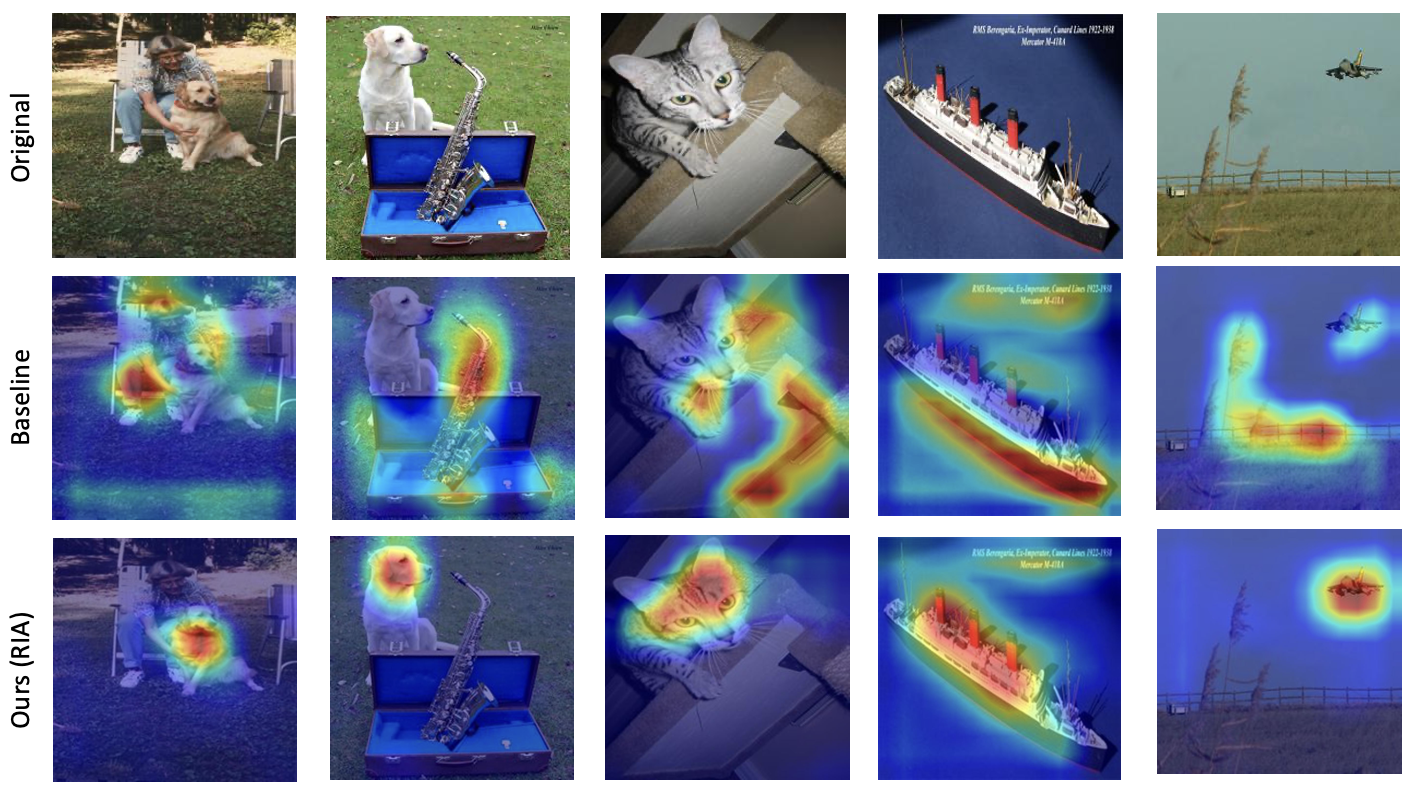}
\caption{Grad-CAM visualization results for images from RIVAL10 validation set using VGG16 model trained from scratch}\label{fig:visualization1}
\vspace{-\baselineskip}
\end{center}
\end{figure}

\begin{figure}[!ht]
\begin{center}
\includegraphics[width=0.8\textwidth]{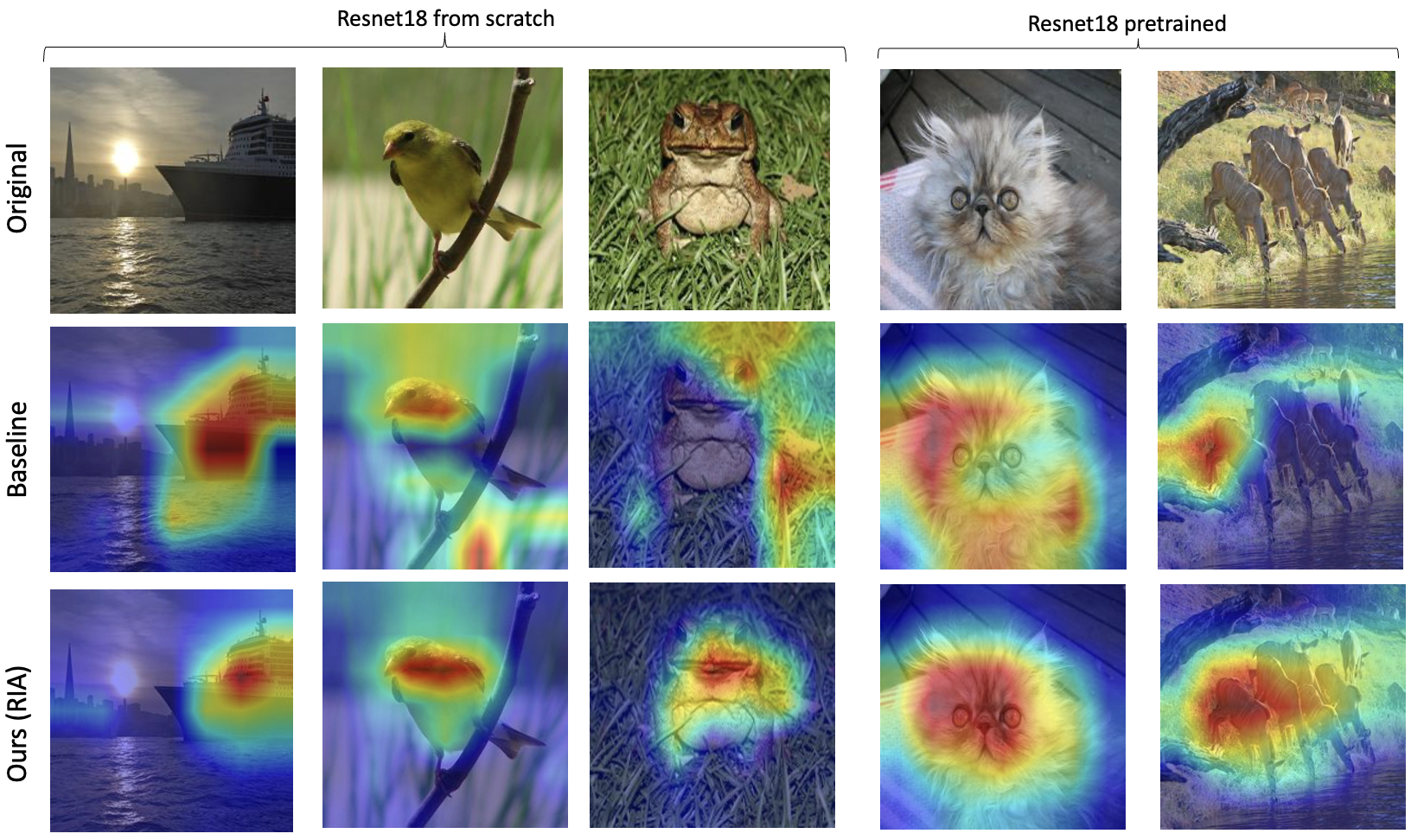}
\caption{Grad-CAM visualization results for images from RIVAL10 validation set using Resnet18}\label{fig:visualization2}
\vspace{-\baselineskip}
\end{center}
\end{figure}

\begin{figure}[!ht]
\begin{center}
\includegraphics[width=0.8\textwidth]{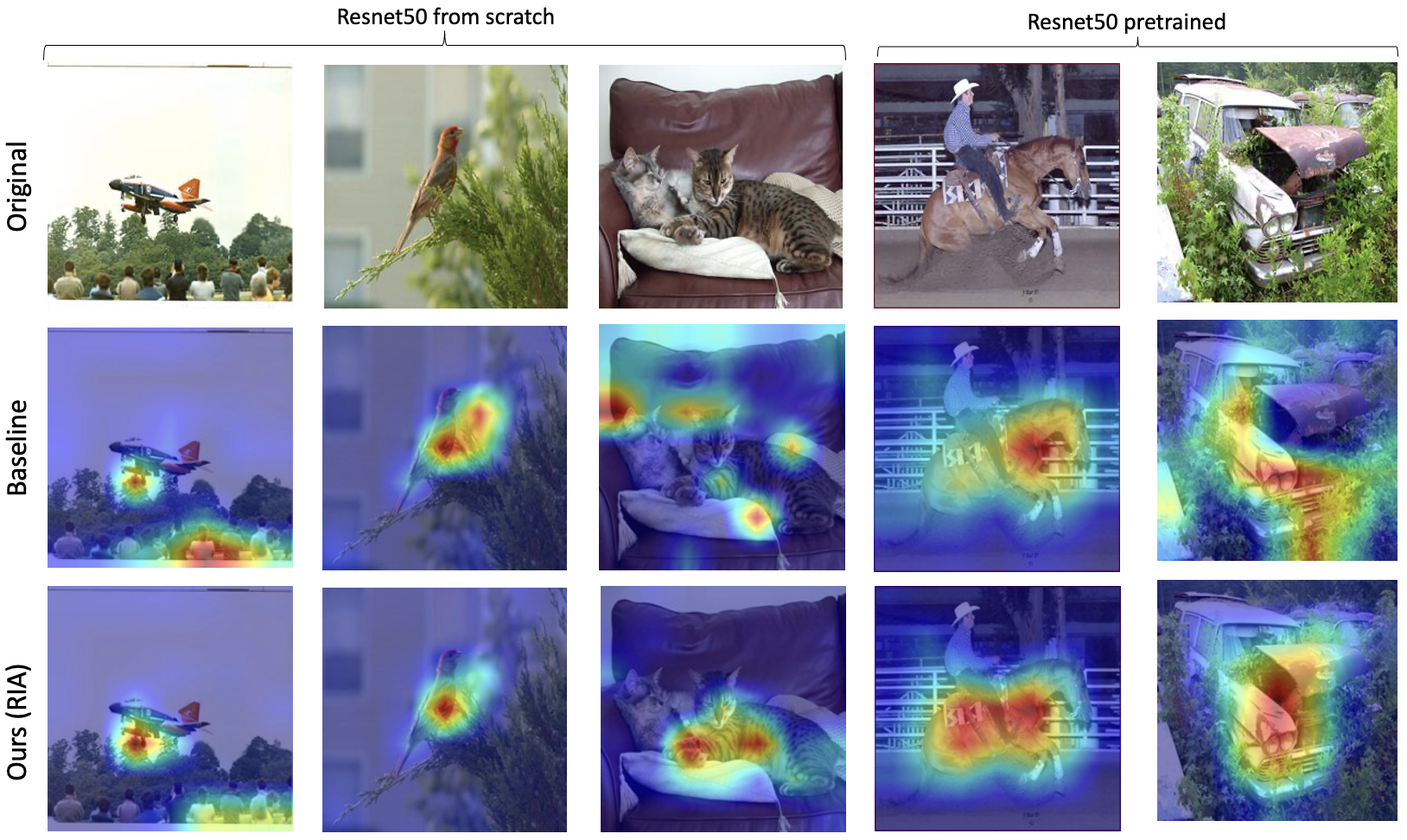}
\caption{Grad-CAM visualization results for images from RIVAL10 validation set using Resnet50}\label{fig:visualization3}
\vspace{-\baselineskip}
\end{center}
\end{figure}

\subsection{Model Explanation}

Our experiments, as shown in Figure \ref{fig:visualization1}, \ref{fig:visualization2}, \ref{fig:visualization3}, have shown that our proposed approach of encouraging the model to focus on the primary object leads to a significant improvement in the interpretability and clarity of the model's decision-making process. By prioritizing the primary object, the model can extract more relevant features and make informed decisions, which is crucial in various computer vision applications. Moreover, our method has demonstrated the ability to reduce bias towards subsidiary objects and concentrate on the desired object even when it is not discernible enough for model. As a result, our approach eliminates bias towards environmental factors, allowing the model to make accurate decisions regardless of the object's location and the presence of other objects. For instance, the model is no longer influenced by water when the object is in water, or by branches when identifying birds, resulting in more robust and reliable explanations.

This improvement is significant as it enhances the reliability and robustness of the model, making it more suitable for real-world scenarios where the reliability of the model is critical, and a wrong decision can have severe consequences. Additionally, the model is trustworthy and  no longer affected by environmental factors, which ensures that it can make accurate decisions regardless of the object's location or the presence of other objects.

\section{Conclusion}
We propose Rigion of Interest Activation (RIA), a novel learning approach that improves the interpretability of deep neural networks by encouraging the model to focus on the primary object's area as much as feasible. We emphasize the importance of evaluating the network based on its quality of explanation, and not only classification accuracy. Our RIA method significantly improves the explanation heatmaps while achieving comparable classification accuracy on RIVAL10 dataset. Additionally, our method can enhance the robustness under foreground and background noises while improving the explanation heatmaps and making the model trustworthy. This demonstrates that our method acts as a regularizer that focuses more attention on the discriminating aspects of the image.






\bibliography{acml23}






\end{document}